\documentclass[5pt]{article}

\usepackage[letterpaper]{geometry}
\usepackage{spconf,amsmath,epsfig}
\usepackage{enumitem}

\usepackage{graphicx}
\usepackage{caption}
\usepackage{multirow}
\usepackage{adjustbox}
\usepackage[colorlinks=true, citecolor={blue}, bookmarks=false]{hyperref}

\usepackage{scrextend}
\changefontsizes{9.5pt}

\usepackage{fancyhdr}
\newcommand{\argmin}{\mathop{\rm argmin}\limits}

\begin{document}\sloppy
\topmargin=0mm

\title{Transformation on Computer--Generated Facial Image to Avoid Detection by Spoofing Detector}

\name{Huy H. Nguyen\textsuperscript{1}, Ngoc-Dung T. Tieu\textsuperscript{1}, Hoang-Quoc Nguyen-Son\textsuperscript{2}, Junichi Yamagishi\textsuperscript{1,2,3}, Isao Echizen\textsuperscript{1,2}}

\address{\textsuperscript{1}Graduate University for Advanced Studies, Kanagawa, Japan\\
	\textsuperscript{2}National Institute of Informatics, Tokyo, Japan\\
	\textsuperscript{3}The University of Edinburgh, Edinburgh, UK\\
	\{nhhuy, dungtieu, nshquoc, jyamagishi, iechizen\}@nii.ac.jp
}

\maketitle
	
\begin{abstract}
	Making computer-generated (CG) images more difficult to detect is an interesting problem in computer graphics and security. While most approaches focus on the image rendering phase, this paper presents a method based on increasing the naturalness of CG facial images from the perspective of spoofing detectors. The proposed method is implemented using a convolutional neural network (CNN) comprising two autoencoders and a transformer and is trained using a black-box discriminator without gradient information. Over 50\% of the transformed CG images were not detected by three state-of-the-art spoofing detectors. This capability raises an alarm regarding the reliability of facial authentication systems, which are becoming widely used in daily life.
\end{abstract}


\section{Introduction}
A presentation attack is commonly used to bypass authentication systems using biometrics information (face, fingerprint, iris, and/or voice). Integration of a spoofing detector into the system before the authentication phase is one approach to preventing such attacks. A good candidate for this is liveness detection, which generally uses a challenge-response protocol in which the user is asked to perform an action such as blinking, smiling, or moving the lips. However, recent work has shown that it is possible to avoid liveness detection by, for example, using real-time face capture and reenactment~\cite{thies2016face2face}. It has thus become necessary to develop and implement natural--CG image/video discriminators.

Forensic research on discriminating between CG and natural images has focused on both images in general and facial images. As an example of the former, Wu et al. extracted statistical features from histograms of differential images~\cite{wu2011identifying}. Although this approach was proposed several years ago, our evaluation demonstrated that it has fast feature extraction and good performance. Therefore, we used it as the basis of the discriminator used to train our CNN. In 2017, Peng et al.~\cite{peng2017discrimination} reported a method based on multi-fractal and regression analysis. As an example of focusing on facial images, the recent work of Nguyen et al.~\cite{nguyen2015discriminating} focused on facial smoothness as represented by edges and local entropy of the skin areas.

\begin{figure}[t]
	\begin{center}
		\includegraphics[width=86mm]{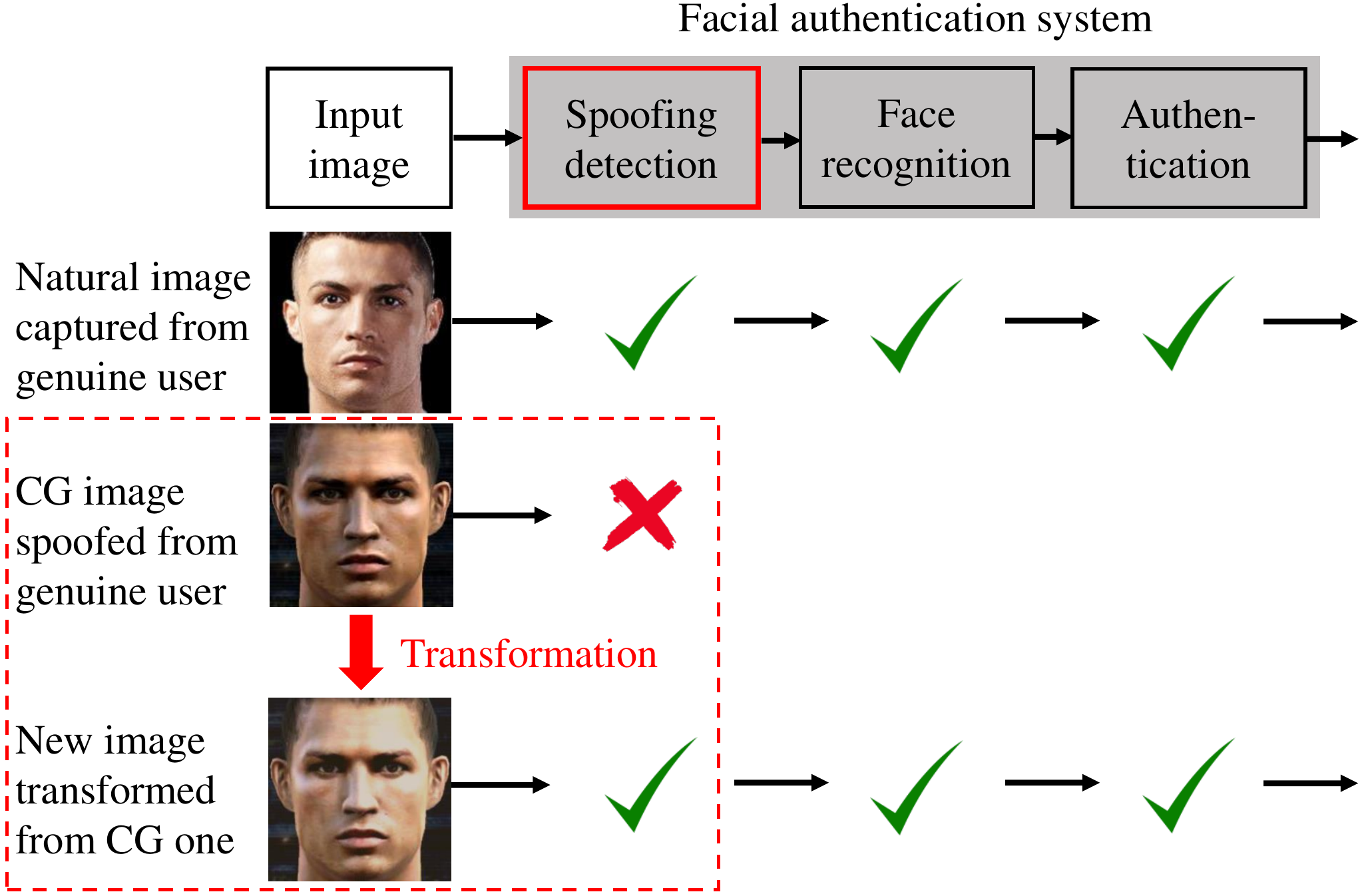}
		\caption{Scenario of proposed presentation attack method.}
		\label{figure:presentation-attack}
	\end{center}
\end{figure}

In the research reported here, we developed a method for avoiding detection by spoofing detectors like those  natural--CG image discriminators mentioned above. It works by transforming CG images input to facial authentication systems to make them appear more natural. Our work is motivated by the idea of ``adversarial machine learning'' of Huang et al. for attacking machine learning based systems~\cite{huang2011adversarial}. Although we did not target a specific system, we used a spoofing detection algorithm as the basis of the discriminator. The attack scenario is illustrated in Fig.~\ref{figure:presentation-attack}. Given two sets of data (a set of natural images and a set of CG ones which are not necessarily corresponding person-to-person or pose-to-pose), a system using the proposed method implemented as a CNN transforms the CG input images in an attempt to make them indistinguishable from the natural counterparts.

Unlike the original generative adversarial network (GAN)~\cite{goodfellow2014generative}, the discriminator used to train our CNN was pre-trained, kept fixed during training, and could not perform back propagation. Moreover, the generated images retained important information from the input images, such as the person’s identity, the facial expression, or the lips' shapes, which are hard to control when using the original GAN. When designing our CNN we focused on minimizing the number of parameters so that it can work without consuming a large amount of GPU memory. We also considered the ``training with small dataset problem,'' which is often faced by attackers when collecting data.

Our approach is different from those of other computer graphic engines, which mainly focus on the rendering phase. It also differs from the style transfer problem, which mainly focuses on applying the ``look'' or textures of one image (the style) to another one (the content). In our case, the style is not just a picture or an artistic style, so it is hard to define.

Our contributions here are threefold:
\begin{itemize}  
	\item Presentation of a CNN comprising two autoencoders and a transformer net that increases the difficulty of detecting computer-generated facial images.
	\item Suggestion of a method for dealing with back propagation when training using an external black-box discriminator that does not have gradient information.
	\item Raising of an alarm about the robustness of facial authentication systems, which are being implemented in many mobile devices and have become a tempting target for attacker.
\end{itemize}

\section{Model Architecture}
\subsection{Overview}
It is very hard to explicitly point out what makes natural images look ``natural'' and CG ones look ``unnatural.'' We humans have been ``trained'' by our encounters with many natural things and scenes since we were born, so we can intuitively distinguish which images were produced by computer. Some forensic researchers have tried describing these intuitive feelings in terms of specific properties~\cite{peng2017discrimination,nguyen2015discriminating}. Others, such as Wu et al.~\cite{wu2011identifying}, have tried using statistical methods to distinguish natural images from CG images. However, these approaches are problematic, require lengthy experiments and do not sufficiently describe the essence of the two kinds of images.

\begin{figure}[t]
	\begin{center}
		\includegraphics[width=62mm]{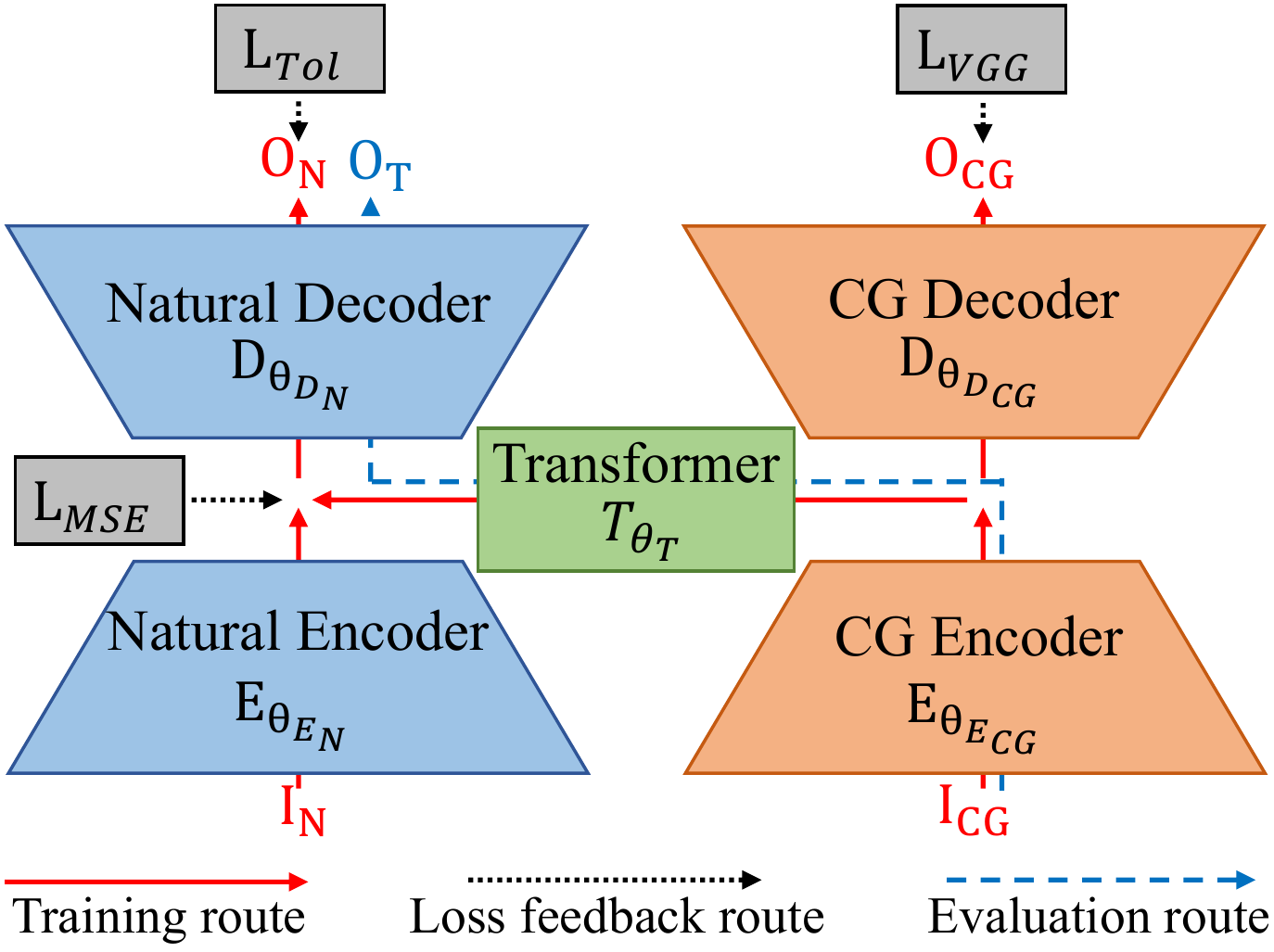}
		\caption{H-Net architecture.}
		\label{figure:h_net}
	\end{center}
\end{figure}

To tackle this feature-finding problem, we use two autoencoders~\cite{bengio2009learning} with the same design but different weights ($\{E_{\theta_{E_N}}, D_{\theta_{D_N}}\}$ and $\{E_{\theta_{E_{CG}}}, D_{\theta_{D_{CG}}}\}$) to automatically extract latent features from natural image $I_N$ and CG image $I_{CG}$. They are trained to learn representations of the inputs and to regenerate them with the same properties. To transform the latent feature space of CG images into those of natural ones, we use a transformer net $T_{\theta_T}$ with five bottleneck residual blocks~\cite{he2016deep}. This block design reduces the number of parameters for the network. Old information given by the skip connection combines with new one created by the layers in a residual block are suitable for transformation. Since our network is shaped like an ``H,'' as illustrated in Fig.~\ref{figure:h_net}, we call it ``H-Net.'' In the evaluation phase, the CG encoder $E_{\theta_{E_{CG}}}$, the transformer $T_{\theta_T}$, and the natural decoder $D_{\theta_{E_N}}$ transform CG input $I_{CG}$ into a natural-looking image $O_T$ that is similar to natural image $I_N$ from the perspective of a natural--CG image discriminator.

\begin{figure}[t]
	\begin{center}
		\includegraphics[width=65mm]{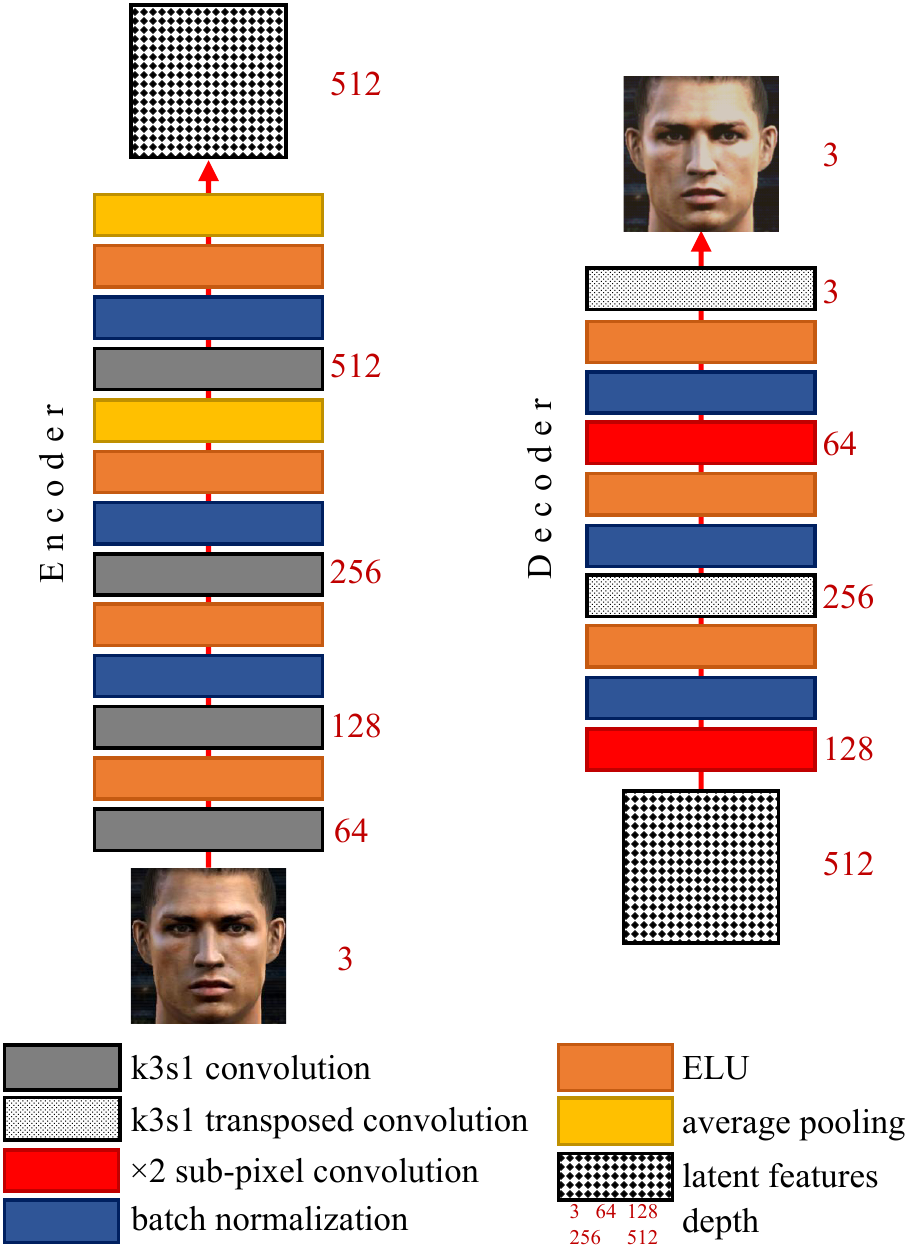}
		\caption{Design of encoder net (left) and decoder net (right).}
		\label{figure:en_de}
	\end{center}
\end{figure}

The encoder has four convolutional layers with $3 \times 3$ kernels and a stride of 1. Batch normalization~\cite{ioffe2015batch} is used to deal with high learning rates and the less careful initialization problem. After batch normalization layers are exponential linear unit (ELU) layers, which give better performance than traditional rectified linear units (ReLUs), leaky ReLUs (LReLUs), and parametrized ReLUs (PReLUs)~\cite{clevert2015fast}. To reduce the spatial size of the representation, we use $2 \times 2$ average pooling layers, which have recently come to be used instead of max pooling ones. In the decoder, the convolutional layers are replaced with their counterparts: transposed convolutional ones and two sub-pixel convolutional ones, which have been reported to give better upscaling results than previous approaches\cite{shi2016real}. The detailed designs of the encoder and decoder are shown in Fig ~\ref{figure:en_de}.

\subsection{Loss functions}
Loss functions are used to optimize the model in the training phase. Pixel-wise loss functions such as mean squared error (MSE) ones are widely used but have limited capabilities for images when measuring perceptual quality~\cite{mathieu2015deep}. To overcome this limitation, Dosovitskiy and Brox~\cite{dosovitskiy2016generating} used Euclidean-based loss in feature space in combination with adversarial loss. Another approach is to use a perceptual loss function based on the Euclidean distance between feature maps extracted from a VGG network proposed by the Visual Geometry Group at the University of Oxford~\cite{simonyan2014very}. Ledig et al. combined both VGG loss and adversarial loss in their super-resolution generative adversarial network~\cite{ledig2016photo}.

We use four loss functions:
\begin{enumerate}
	\item MSE loss $L_{MSE}$ is only used to calculate the loss between two feature maps, including high-level presentations encoded by the encoders and others extracted from the VGG-19 network.
	
	\item VGG loss $L_{VGG}$, or perceptual loss, is the MSE loss of features maps (of $I_1$ and $I_2$) extracted using the VGG-19 network: $L_{VGG} = L_{MSE}(VGG(I_1), VGG(I_2))$. VGG loss represents the perceptual quality of generated images.
	
	\item Adversarial loss $L_{Adv}$ is the binary cross entropy loss between two labels: one is from the discriminator and the other is the destination label. Details of the discriminator are discussed in section~\ref{sec:discriminator}.
	
	\item Total loss $L_{Tol}$ is the combination of $L_{VGG}$ and $L_{Adv}$, formulated as equation~\ref{eq:total_loss}. It represents the quality of images generated by natural decoder $D_{\theta_{E_N}}$. In our experiments, we set the hyper-parameter $\alpha$ to $5 \times 10^{-3}$. Gradient back propagation is discussed in section~\ref{sec:discriminator}.
	\begin{equation}
		L_{Tol} = (1 - \alpha)L_{VGG} + \alpha L_{Adv}
		\label{eq:total_loss}
	\end{equation}
\end{enumerate}

\subsection{Discriminator} 
\label{sec:discriminator}
Unlike the original GAN proposed by Goodfellow et al.~\cite{goodfellow2014generative}, we use a pre-trained discriminator for training our H-Net. It can be any black-box spoofing detector. Since it is unlikely for attackers in the wild to have full access to the spoofing detectors in authentication systems, it is more realistic to assume that the spoofing detector used as the discriminator used for training is a ``black-box'' rather than a differentiable white-box discriminator. In our experiments, the use of a ``traditional'' discriminator in the GAN, such as the one used by Ledig et al.~\cite{ledig2016photo}, results in very poor performance. This is because performing only convolution is not enough for extracting the features needed for distinguishing natural images from CG images. Hence, we borrow the method of Wu et al., which focuses on statistical features from histograms of differential images~\cite{wu2011identifying}. We use a neural net instead of Fisher's linear discriminant analysis (FLDA) as Wu et al. did because we can easily run the discriminator in parallel with H-Net in the training phase without using an additional framework. 

We need to compute the gradient of total loss $L_{Tol}$ in equation~\ref{eq:total_loss} with respect to the parameters in the generator. However, since the discriminator is a black-box, the adversarial loss $L_{Adv}$ is not differentiable. One possible approach is to approximate it on the basis of the gradient of the VGG loss $L_{VGG}$ only. However, the adversarial loss makes no contribution to the gradient. Therefore, we approximate the total gradient as follows:

\begin{equation}
	\begin{aligned}
		grad_{L_{Tol}} = \frac{(1 - \alpha)L_{VGG} + \alpha  L_{Adv}} {L_{VGG}} grad_{L_{VGG}}
	\end{aligned}
	\label{eq:grad_total}
\end{equation}

\subsection{Training and evaluation}
H-Net is trained using two sets of images: natural images and CG ones. The two sets do not have to be pairwise correlated, which means that they may contains images of different people with different poses or facial expressions. This loose condition comes from reality; i.e., it is very hard for an attacker to find a large number of natural--CG image pairs with similar perceptual content.

In training, four optimization processes are performed sequentially:

\textbf{Step 1:} Training the natural autoencoder $\{E_{\theta_{E_N}}, D_{\theta_{D_N}}\}$ with natural image $I_N$ by minimizing equation~\ref{eq:hnet_1}.
\begin{equation}
	\begin{aligned}
		\argmin_{\theta_{E_N},\theta_{D_N}}L_{Tol}[D_{\theta_{D_N}}(E_{\theta_{E_N}}(I_{N})), I_{N}]
		\label{eq:hnet_1}
	\end{aligned}
\end{equation}
The output image $O_N = D_{\theta_{D_N}}(E_{\theta_{E_N}}(I_{N}))$ must satisfy two conditions: (1) have the same perceptual content as input $I_N$ and (2) be classified as a natural image by the discriminator.

\textbf{Step 2:} Training the CG autoencoder $\{E_{\theta_{E_{CG}}}, D_{\theta_{D_{CG}}}\}$ with CG image $I_{CG}$ by minimizing equation~\ref{eq:hnet_2}.
\begin{equation}
	\begin{aligned}
		\argmin_{\theta_{E_{CG}},\theta_{D_{CG}}}L_{VGG}[D_{\theta_{D_{CG}}}(E_{\theta_{E_{CG}}}(I_{CG})), I_{CG}]
		\label{eq:hnet_2}
	\end{aligned}
\end{equation}
In this step, the discriminator loss is not necessary. The only requirement is that output image $O_{CG}$ be perceptually similar to $I_{CG}$.

\textbf{Step 3:} Training the transformer net $T_{\theta_T}$ to convert latent features encoded by CG encoder $E_{\theta_{E_{CG}}}$ into ones that have the same distribution of features as those extracted by the natural encoder $E_{\theta_{E_N}}$. As mentioned above, due to the lack of pairwise correlated $I_N$ and $I_{CG}$ pairs, it is very hard to minimize the loss between $E_{\theta_{E_N}}(I_N)$ and $T_{\theta_T}(E_{\theta_{E_{CG}}}(I_{CG}))$. An acceptable solution is feeding $I_N$ into both networks and minimizing the features they encode, formulated in equation~\ref{eq:hnet_3}.
\begin{equation}
	\begin{aligned}
		\argmin_{\theta_T}L_{MSE}[E_{\theta_{E_N}}(I_{N}),T_{\theta_T}(E_{\theta_{E_{CG}}}(I_N))]
		\label{eq:hnet_3}
	\end{aligned}
\end{equation}

\textbf{Step 4:} Training the CG transformation path, which includes CG encoder $E_{\theta_{E_{CG}}}$, transformer $T_{\theta_T}$, and natural decoder $D_{\theta_{D_N}}$. As in the first step, total loss must be used to ensure that transformed output $O_T = D_{\theta_{D_N}}(T_{\theta_T}(E_{\theta_{E_{CG}}}(I_{CG})))$ is not classified as a CG image. The lack of pairwise correlated $I_N$ and $I_{CG}$ pairs in Step 3 appears here as well, so we need to replace $I_N$ with $I_{CG}$ in equation~\ref{eq:hnet_4}.
\begin{equation}
	\begin{aligned}
		\argmin_{\theta_{E_{CG}},\theta_T,\theta_{D_N}}L_{Tol}[D_{\theta_{D_N}}(T_{\theta_T}(E_{\theta_{E_{CG}}}(I_{CG}))), I_{CG}]
		\label{eq:hnet_4}
	\end{aligned}
\end{equation}
Doing this could also bring back some ``balance'' from the effect of $I_{N}$ on CG encoder $E_{\theta_{E_{CG}}}$ in step 3.

When we compute the transformed output from the CG input in the evaluation phase, we use a step similar to step 4 in the training phase:
\begin{equation}
	O_T = D_{\theta_{D_N}}(T_{\theta_T}(E_{\theta_{E_{CG}}}(I_{CG})))
	\label{eq:hnet_transform}
\end{equation}
We do not use the $tanh$ function to scale the output images because of the resulting low contrast. To avoid extreme values with resulting abnormal color spots, the pixel intensities are restricted to the range $[-1.8,1.8]$ before being denormalized to $[0,255]$.

\section{Evaluation}
In facial image forensic research, there was no standardized dataset used for mutual comparison. Each research group tended to have its own datasets. Therefore, we had to combine pieces from five different sources to create three datasets used for evaluation (see Table~\ref{tab:dataset}). All images were resized to 256 $\times$ 256 pixel resolution. We tested H-Net on two scenarios: (1) the attacker knows the dataset used for training the spoofing detector, and (2) the attacker has no knowledge of the training dataset. Three spoofing detectors were used: Wu et al.'s~\cite{wu2011identifying}, Peng et al.'s~\cite{peng2017discrimination}, and Nguyen et al.'s~\cite{nguyen2015discriminating}. Note that only Wu et al.'s spoofing detector was used as a discriminator for training H-Net; the other spoofing detectors were unseen by the attacker. Moreover, in Wu et al.'s one, we used FLDA for classification as in the authors' report. We compared both the accuracies and detection rates of the three spoofing detectors. Let $n_{TP}$ and $n_{TN}$ be respectively the number of images correctly classified as CG or natural, $n_{FN}$ be the number of CG or transformed images misclassified as natural ones, and $N$ be the total number of images. Accuracy is defined as sum of $n_{TP}$ and $n_{TN}$ over $N$: $\frac{n_{TP} + n_{TN}}{N}$. The detection rate represents the ability of the spoofing detector to detect positive items: $\frac{n_{TP}}{n_{TP} + n_{FN}}$. 

\begin{table*}[t!]
	\centering
	\caption{Three datasets used in evaluation.}
	\label{tab:dataset}
	\begin{tabular}{|c|l|l|l|}
		\hline
		\textbf{No.} & \multicolumn{1}{c|}{\textbf{Components}} & \multicolumn{1}{c|}{\textbf{Size}} & \multicolumn{1}{c|}{\textbf{Description}} \\ \hline
		1 & Dang-Nguyen et al.~\cite{dang2012discrimination} & \begin{tabular}[c]{@{}l@{}}CG: 240\\ \\ Nat: 240\end{tabular} & \begin{tabular}[c]{@{}l@{}}40 very realistic CG images collected from Web plus\\ 200 good quality CG images extracted from PES 2012 soccer game\\ 240 natural images retrieved from Internet\end{tabular} \\ \hline
		2 & \begin{tabular}[c]{@{}l@{}}Basel (CG)~\cite{paysan20093d}\\ Caltech99 (Nat)~\cite{weber1999caltech}\end{tabular} & \begin{tabular}[c]{@{}l@{}}CG: 270\\ Nat: 270\end{tabular} & \begin{tabular}[c]{@{}l@{}}3D face scans and rendered images from Basel Face Model\\ Natural images from Caltech Faces 1999 dataset\end{tabular} \\ \hline
		3 & \begin{tabular}[c]{@{}l@{}}MIT (CG - Grayscale)~\cite{weyrauch2004component}\\ MS-Celeb-1M (Nat)~\cite{guo2016ms}\end{tabular} & \begin{tabular}[c]{@{}l@{}}CG: 3236\\ Nat: 3236\end{tabular} & \begin{tabular}[c]{@{}l@{}}CG images extracted from MIT CBCL dataset\\ Natural images selected from MS-Celeb-1M cropped version\end{tabular} \\ \hline
	\end{tabular}
\end{table*}

\begin{table*}[th!]
	\centering
	\caption{Scenario 1 - Accuracies (\%) and detection rates (\%) of three spoofing detectors on three datasets before and after performing transformation on CG parts.}
	\label{tab:exp_1}
	\begin{adjustbox}{width=1\textwidth}
		\begin{tabular}{|l|r|r|r|r|r|r|r|r|r|r|r|r|}
			\hline
			\multirow{3}{*}{\textbf{Spoofing detectors}} & \multicolumn{4}{c|}{\textbf{Dataset 1}} & \multicolumn{4}{c|}{\textbf{Dataset 2}} & \multicolumn{4}{c|}{\textbf{Dataset 3}} \\ \cline{2-13} 
			& \multicolumn{2}{c|}{\textbf{Accuracy}} & \multicolumn{2}{c|}{\textbf{Detection rate}} & \multicolumn{2}{c|}{\textbf{Accuracy}} & \multicolumn{2}{c|}{\textbf{Detection rate}} & \multicolumn{2}{c|}{\textbf{Accuracy}} & \multicolumn{2}{c|}{\textbf{Detection rate}} \\ \cline{2-13} 
			& \textbf{Before} & \textbf{After$\downarrow$} & \textbf{Before} & \textbf{After$\downarrow$} & \textbf{Before} & \textbf{After$\downarrow$} & \textbf{Before} & \textbf{After$\downarrow$} & \textbf{Before} & \textbf{After$\downarrow$} & \textbf{Before} & \textbf{After$\downarrow$} \\ \hline
			Wu et al.~\cite{wu2011identifying} & 92.71 & \textbf{48.33} & 93.33 & \textbf{0.00} & 83.89 & \textbf{55.37} & 93.33 & \textbf{36.30} & 64.65 & \textbf{56.83} & 35.51 & \textbf{19.90} \\
			Peng et al.~\cite{peng2017discrimination} & 90.63 & \textbf{52.71} & 92.08 & \textbf{16.25} & 59.26 & \textbf{19.26} & 97.41 & \textbf{17.41} & 50.20 & \textbf{13.58} & 100.00 & \textbf{26.79} \\
			Nguyen et al.~\cite{nguyen2015discriminating} & 83.54 & \textbf{64.79} & 87.08 & \textbf{48.33} & \textbf{17.78} & 32.78 & \textbf{0.00} & 30.00 & \textbf{32.31} & 63.23 & \textbf{0.49} & 62.11 \\ \hline
		\end{tabular}
	\end{adjustbox}
\end{table*}

\begin{figure}[th!]
	\begin{center}
		\includegraphics[width=86mm]{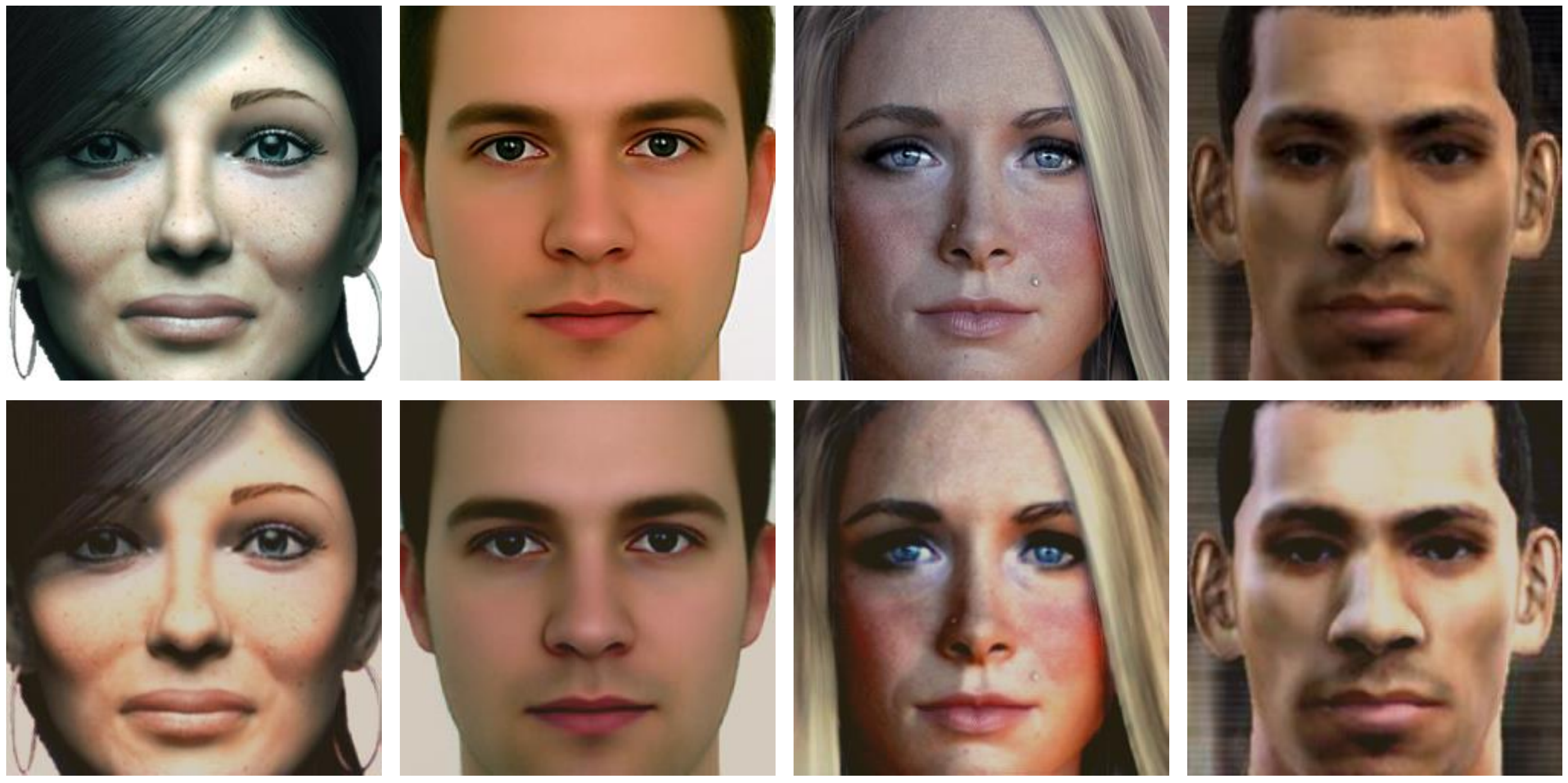}
		\caption{Original images (top) and transformed ones (bottom) from dataset 1. Color and brightness of images in second row were normalized by H-Net as learned from training data. First two images demonstrate good transformation in perception. Contrast of third image was improved. In last image, skin color was whitened a bit undesirably.}
		\label{figure:dataset_1}
	\end{center}
\end{figure}

\begin{figure}[th!]
	\begin{center}
		\includegraphics[width=86mm]{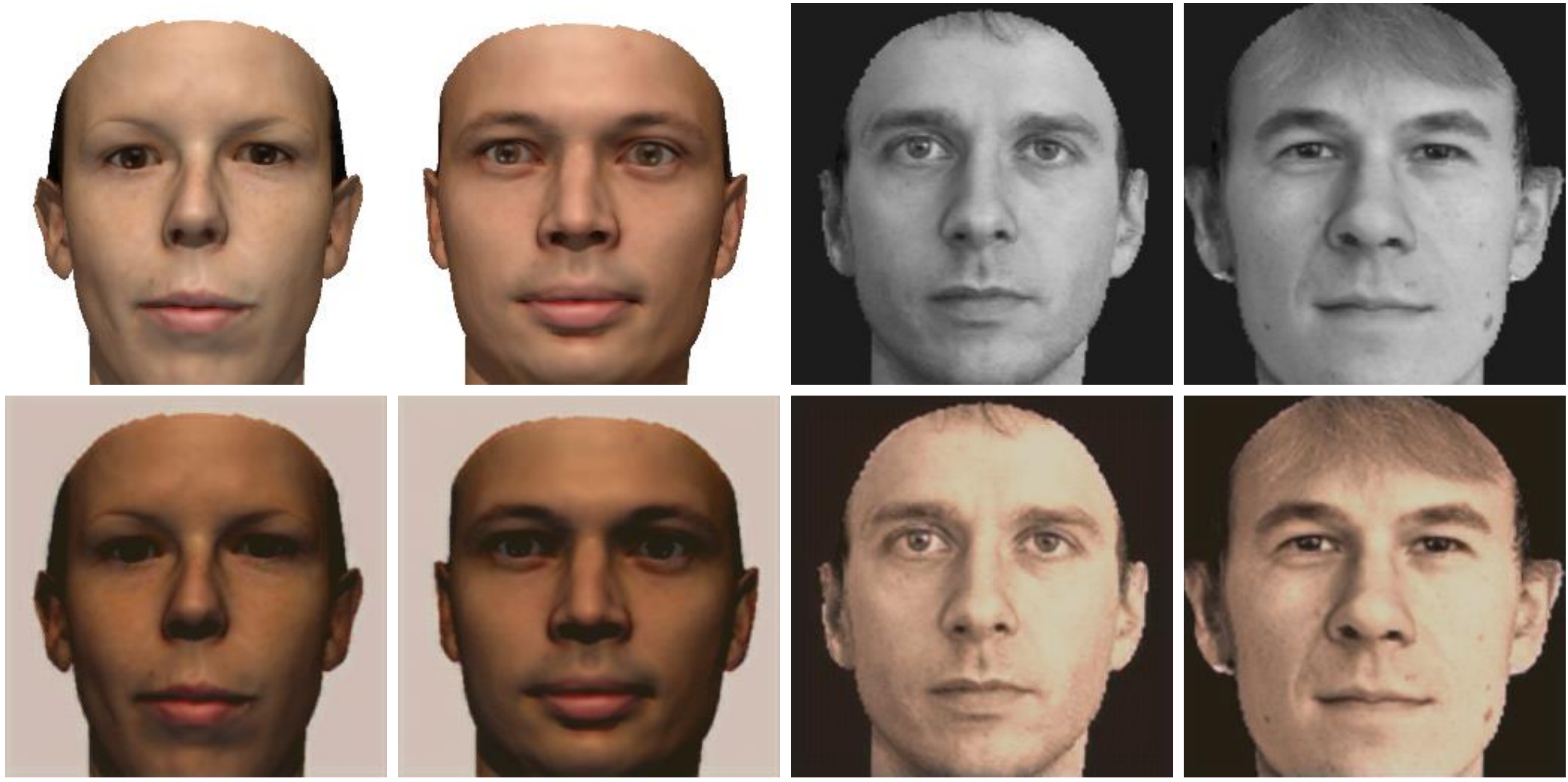}
		\caption{Original images (top) and transformed ones (bottom) from dataset 2 (left) and dataset 3 (right). Brightness of first two image was unexpectedly reduced due to bright background. Although grayscale images were given skin-like color, they can easily be converted back into grayscale.\\
			\copyright Copyright University of Basel. \copyright Copyright 2003-2005 Massachusetts Institute of Technology. All Right Reserved.\\}
		\label{figure:dataset_23}
	\end{center}
\end{figure}

\subsection{Scenario 1: Attacker knows training dataset of spoofing detector}
We trained both spoofing detectors and H-Net on dataset 1. We then evaluated them on all datasets to see if the images transformed by H-Net could avoid detection by these pre-trained spoofing detectors. Comparisons of sample images before and after transformation are shown in Figs.~\ref{figure:dataset_1} and~\ref{figure:dataset_23}. As shown in Table~\ref{tab:exp_1}, the detection rates significantly decreased when the CG images were transformed, especially for Wu et al.’s and Peng et al.'s methods.

Although the method of Nguyen et al. had the lowest performance on dataset 1, its detection rate after transformation was the highest (nearly 50\%). On datasets 2 and 3, its performance before transformation was very poor; it was better after transformation. Our analysis shows that this method was over-fitted for dataset 1 which had high-quality CG images. It had a tendency to classify fine-texture images as CG ones. On datasets 2 and 3, which did not have fine-texture CG images, it classified almost of the images as natural ones. Because of the spoofing detector's preset threshold, some transformed images had good enough texture to be classified as CG ones, which increased the detection rate. Therefore, if an attacker tries to avoid detection by this spoofing detector, he or she may be successful the first time without the help of H-Net.

\subsection{Scenario 2: Attacker does not know training dataset of spoofing detector}
Unlike in the first scenario, we trained the H-Net and the spoofing detectors on difference datasets (switched between dataset 1 and 2), and evaluated them on dataset 3.
\subsubsection{Scenario 2.1: H-Net was trained on dataset 1, spoofing detectors were trained on dataset 2}
The evaluation results on dataset 3 are shown in Table~\ref{tab:exp_21}. The transformed images again significantly reduced the detection rates of all spoofing detectors, especially those of Wu et al. and Peng et al., which were nearly 0\%. Nguyen et al.'s method had a detection rate of around 50\%, down from nearly 100\%, meaning that the attacker had a 50--50 chance of avoiding detection by this spoofing detector. In this case, this spoofing detector learned that low-texture images had a high probability of being CG ones, which was opposite to its knowledge in scenario 1. Therefore, after transformation on dataset 3, the textures of the CG images were improved so that the images would likely be classified as natural ones. This also clarifies the contrasting changes in the detection rate in the first scenario on dataset 1 vs. datasets 2 and 3.

\begin{table}[th!]
	\centering
	\caption{Scenario 2.1 - Evaluation results on dataset 3.}
	\label{tab:exp_21}
	\begin{tabular}{|l|c|c|c|c|}
		\hline
		\multirow{2}{*}{\textbf{Spoofing detectors}} & \multicolumn{2}{c|}{\textbf{Accuracy}} & \multicolumn{2}{c|}{\textbf{Detection rate}} \\ \cline{2-5} 
		& \textbf{Before} & \textbf{After} & \textbf{Before} & \textbf{After} \\ \hline
		Wu et al.~\cite{wu2011identifying} & 56.38 & \textbf{6.46} & 100.00 & \textbf{0.19} \\
		Peng et al.~\cite{peng2017discrimination} & 92.32 & \textbf{42.57} & 100.00 & \textbf{0.49} \\
		Nguyen et al.~\cite{nguyen2015discriminating} & 96.72 & \textbf{71.54} & 99.20 & \textbf{48.89} \\ \hline
	\end{tabular}
\end{table}

\subsubsection{Scenario 2.2: H-Net was trained on dataset 2, spoofing detectors were trained on dataset 1}
The evaluation results on dataset 3 are shown in Table~\ref{tab:exp_22}. Before transformation, Peng et al.'s method seemed to classify all input as CG. After transformation, its decision was changed that all transformed CG images were classified as natural ones, therefore both the accuracy and the detection rate were around 0. Nguyen et al.'s method had the same behavior as in the first scenario as expected. The performance of Wu et al.'s method was increased after transformation, which was different from the two above scenarios. A possible explanation for this phenomenon is that the training dataset used for H-Net is small and monotonous. As the result, H-Net did not have enough knowledge about other kinds of CG images.

\begin{table}[th!]
	\centering
	\caption{Scenario 2.2 - Evaluation results on dataset 3.}
	\label{tab:exp_22}
	\begin{tabular}{|l|c|c|c|c|}
		\hline
		\multirow{2}{*}{\textbf{Spoofing detectors}} & \multicolumn{2}{c|}{\textbf{Accuracy}} & \multicolumn{2}{c|}{\textbf{Detection rate}} \\ \cline{2-5} 
		& \textbf{Before} & \textbf{After} & \textbf{Before} & \textbf{After} \\ \hline
		Wu et al.~\cite{wu2011identifying} & \textbf{64.65} & 82.73 & \textbf{35.51} & 71.66 \\
		Peng et al.~\cite{peng2017discrimination} & 50.20 & \textbf{0.20} & 100.00 & \textbf{0.00} \\
		Nguyen et al.~\cite{nguyen2015discriminating} & \textbf{32.31} & 41.27 & \textbf{0.49} & 18.17 \\ \hline
	\end{tabular}
\end{table}

\section{Conclusion and Future Work}
The performances of both H-Net and detectors are depended on the quality of training datasets. However, in most cases, over 50\% of the CG images transformed using our H-Net avoided detection by three state-of-the-art spoofing detectors. Since the facial features were preserved, facial recognition was unaffected. This means that the network can be trained using a black-box discriminator that cannot perform back propagation. However, H-Net has some limitations, especially when up to 50\% transformed images are still separable from natural ones. Future work will mainly focus on this limitation as well as finding better datasets. We will also evaluate the local substitute method to perform black-box attacks~\cite{papernot2017practical}. Other issues to be addressed are solving the black skin-color problem, dealing with larger images, and reducing network size to enable it to work smoothly with video frames in real time.

\section*{Acknowledgments}
This work was supported by JSPS KAKENHI Grant Numbers JP16H06302 and
15H01686.

We are grateful to the Massachusetts Institute of Technology and to the Center for Biological and Computational Learning for providing databases of facial images. We thank Prof. Dr. Thomas Vetter, Department of Computer Science at the University of Basel, for the 3D face scans and renderings dataset. We also thank the Computer Vision Department of Caltech University, Microsoft Corporation, Dr. Duc-Tien Dang-Nguyen and his colleges for providing invaluable facial datasets for conducting our experiments. 

\bibliographystyle{IEEEbib}
\bibliography{nhhuy}

\begin{thebibliography}{10}

\bibitem{thies2016face2face}
Justus Thies, Michael Zollhofer, Marc Stamminger, Christian Theobalt, and
  Matthias Nie{\ss}ner,
\newblock ``{Face2Face}: Real-time face capture and reenactment of {RGB}
  videos,''
\newblock in {\em CVPR}, 2016, pp. 2387--2395.

\bibitem{wu2011identifying}
Ruoyu Wu, Xiaolong Li, and Bin Yang,
\newblock ``Identifying computer generated graphics via histogram features,''
\newblock in {\em ICIP}. IEEE, 2011, pp. 1933--1936.

\bibitem{peng2017discrimination}
Fei Peng, Die-lan Zhou, Min Long, and Xing-ming Sun,
\newblock ``Discrimination of natural images and computer generated graphics
  based on multi-fractal and regression analysis,''
\newblock {\em AEU-International Journal of Electronics and Communications},
  vol. 71, pp. 72--81, 2017.

\bibitem{nguyen2015discriminating}
Huy~H Nguyen, Hoang-Quoc Nguyen-Son, Thuc~D Nguyen, and Isao Echizen,
\newblock ``Discriminating between computer-generated facial images and natural
  ones using smoothness property and local entropy,''
\newblock in {\em IWDW}. Springer, 2015, pp. 39--50.

\bibitem{huang2011adversarial}
Ling Huang, Anthony~D Joseph, Blaine Nelson, Benjamin~IP Rubinstein, and
  JD~Tygar,
\newblock ``Adversarial machine learning,''
\newblock in {\em ACM Workshop on Security and Artificial Intelligence}. ACM,
  2011, pp. 43--58.

\bibitem{goodfellow2014generative}
Ian Goodfellow, Jean Pouget-Abadie, Mehdi Mirza, Bing Xu, David Warde-Farley,
  Sherjil Ozair, Aaron Courville, and Yoshua Bengio,
\newblock ``Generative adversarial nets,''
\newblock in {\em NIPS}, 2014, pp. 2672--2680.

\bibitem{bengio2009learning}
Yoshua Bengio et~al.,
\newblock ``Learning deep architectures for {AI},''
\newblock {\em Foundations and trends{\textregistered} in Machine Learning},
  vol. 2, no. 1, pp. 1--127, 2009.

\bibitem{he2016deep}
Kaiming He, Xiangyu Zhang, Shaoqing Ren, and Jian Sun,
\newblock ``Deep residual learning for image recognition,''
\newblock in {\em CVPR}, 2016, pp. 770--778.

\bibitem{ioffe2015batch}
Sergey Ioffe and Christian Szegedy,
\newblock ``Batch normalization: Accelerating deep network training by reducing
  internal covariate shift,''
\newblock in {\em ICML}, 2015, pp. 448--456.

\bibitem{clevert2015fast}
Djork-Arn{\'e} Clevert, Thomas Unterthiner, and Sepp Hochreiter,
\newblock ``Fast and accurate deep network learning by exponential linear units
  ({ELUs}),''
\newblock {\em ICLR}, 2016.

\bibitem{shi2016real}
Wenzhe Shi, Jose Caballero, Ferenc Husz{\'a}r, Johannes Totz, Andrew~P Aitken,
  Rob Bishop, Daniel Rueckert, and Zehan Wang,
\newblock ``Real-time single image and video super-resolution using an
  efficient sub-pixel convolutional neural network,''
\newblock in {\em CVPR}, 2016, pp. 1874--1883.

\bibitem{mathieu2015deep}
Michael Mathieu, Camille Couprie, and Yann LeCun,
\newblock ``Deep multi-scale video prediction beyond mean square error,''
\newblock {\em ICLR}, 2016.

\bibitem{dosovitskiy2016generating}
Alexey Dosovitskiy and Thomas Brox,
\newblock ``Generating images with perceptual similarity metrics based on deep
  networks,''
\newblock in {\em NIPS}, 2016, pp. 658--666.

\bibitem{simonyan2014very}
Karen Simonyan and Andrew Zisserman,
\newblock ``Very deep convolutional networks for large-scale image
  recognition,''
\newblock {\em ICLR}, 2015.

\bibitem{ledig2016photo}
Christian Ledig, Lucas Theis, Ferenc Husz{\'a}r, Jose Caballero, Andrew
  Cunningham, Alejandro Acosta, Andrew Aitken, Alykhan Tejani, Johannes Totz,
  Zehan Wang, et~al.,
\newblock ``Photo-realistic single image super-resolution using a generative
  adversarial network,''
\newblock {\em Computing Research Repository (CoRR)}, 2016.

\bibitem{dang2012discrimination}
Duc-Tien Dang-Nguyen, Giulia Boato, and Francesco~GB De~Natale,
\newblock ``Discrimination between computer generated and natural human faces
  based on asymmetry information,''
\newblock in {\em EUSIPCO}. IEEE, 2012, pp. 1234--1238.

\bibitem{paysan20093d}
Pascal Paysan, Reinhard Knothe, Brian Amberg, Sami Romdhani, and Thomas Vetter,
\newblock ``A 3{D} face model for pose and illumination invariant face
  recognition,''
\newblock in {\em Advanced Video and Signal Based Surveillance}. IEEE, 2009,
  pp. 296--301.

\bibitem{weber1999caltech}
M~Weber and M~Weber,
\newblock ``Caltech frontal face database,''
\newblock {\em California Institute of Technology}, 1999.

\bibitem{weyrauch2004component}
Benjamin Weyrauch, Bernd Heisele, Jennifer Huang, and Volker Blanz,
\newblock ``Component-based face recognition with 3{D} morphable models,''
\newblock in {\em CVPR Workshop}. IEEE, 2004, pp. 85--85.

\bibitem{guo2016ms}
Yandong Guo, Lei Zhang, Yuxiao Hu, Xiaodong He, and Jianfeng Gao,
\newblock ``M{S}-{C}eleb-1{M}: A dataset and benchmark for large scale face
  recognition,''
\newblock in {\em ECCV}. Springer, 2016, pp. 87--102.

\bibitem{papernot2017practical}
Nicolas Papernot, Patrick McDaniel, Ian Goodfellow, Somesh Jha, Z~Berkay Celik,
  and Ananthram Swami,
\newblock ``Practical black-box attacks against machine learning,''
\newblock in {\em AsiaCCS}. ACM, 2017, pp. 506--519.

\end{thebibliography}

\end{document}